\documentclass[10pt,twocolumn,letterpaper]{article}

\usepackage{iccv}
\usepackage{times}
\usepackage{epsfig}
\usepackage{graphicx}
\usepackage{amsmath}
\usepackage{amssymb}
\usepackage{wasysym}
\usepackage[labelformat=simple]{subfig}
\usepackage{multirow}
\newcommand{\splitcell}[1]{\begin{tabular}{@{}l@{}}#1\end{tabular}}
\newcommand{\bsplitcell}[1]{$\left[\splitcell{#1}\right]$}
\usepackage{graphicx}
\usepackage{wrapfig}
\usepackage{caption} 
\usepackage{booktabs}       

\usepackage{enumitem}
\setlist[itemize]{align=parleft,left=0pt..1em}

\usepackage[pagebackref=true,breaklinks=true,letterpaper=true,colorlinks,bookmarks=false]{hyperref}

\iccvfinalcopy 

\def\iccvPaperID{19} 

\ificcvfinal\fi

\begin{document}

\title{Contextual Convolutional Neural Networks}

\author{Ionut Cosmin Duta$^*$, Mariana Iuliana Georgescu, Radu Tudor Ionescu \\
University of Bucharest, Romania; SecurifAI, Romania\\
{\tt\small $^*$icduta@gmail.com}
}

\maketitle
\ificcvfinal\thispagestyle{empty}\fi

\begin{abstract}
We propose contextual convolution (CoConv) for visual recognition. CoConv is a direct replacement of the standard convolution, which is the core component of convolutional neural networks. CoConv is implicitly equipped with the capability of incorporating contextual information while maintaining a similar number of parameters and computational cost compared to the standard convolution. CoConv is inspired by neuroscience studies indicating that $(i)$ neurons, even from the primary visual cortex (V1 area), are involved in detection of contextual cues and that $(ii)$ the activity of a visual neuron can be influenced by the stimuli placed entirely outside of its theoretical receptive field. On the one hand, we integrate CoConv in the widely-used residual networks and show improved recognition performance over baselines on the core tasks and benchmarks for visual recognition, namely image classification on the ImageNet data set and object detection on the MS COCO data set. On the other hand, we introduce CoConv in the generator of a state-of-the-art Generative Adversarial Network, showing improved generative results on CIFAR-10 and CelebA. Our code is available at {\small \url{https://github.com/iduta/coconv}}.
\end{abstract}

\vspace{-0.2cm}
\section{Introduction}
\begin{figure}[!t]
\captionsetup[subfigure]{labelformat=empty}
\centering
\subfloat[(a)]
{	
	\includegraphics[width=0.3\linewidth]{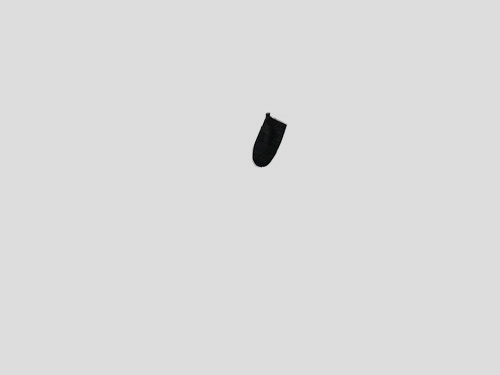}
	\label{Fig_Ambiguous_Glove_subfig1}
}
\subfloat[(b)]
{
	\includegraphics[width=0.3\linewidth]{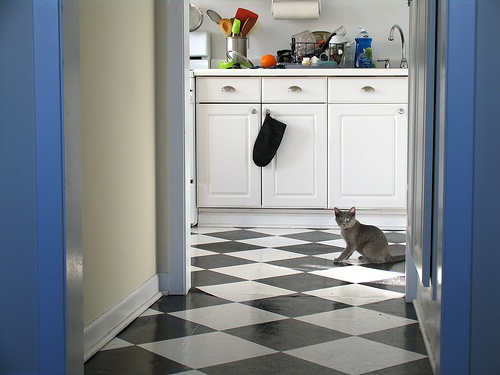}
	\label{Fig_Ambiguous_Glove_subfig2}
}
\vspace{-0.2cm}
\caption{An example in which the context is crucial in labeling the object (kitchen glove), which can easily be mistaken for something else if the rest of the image is not seen. (a) A picture of a kitchen glove. (b) A picture of the same glove with context.\label{fig:context_example}}
\vspace{-0.4cm}
\end{figure}

\vspace{-0.1cm}
Contextual information is vital for a visual perception system. A point (or a small patch) in a scene (image) is mostly meaningless without the surrounding contextual information. For instance, it is very difficult for a person to provide a semantic label or a description for a small patch (taken from an image) without providing a broader visual context. As shown in the example illustrated in Figure~\ref{fig:context_example}, it is even hard to label an entire object without context, let alone some part of the respective object. 
In neuroscience research, the critical role of the contextual influences on visual perception systems is well proven since long time ago~\cite{albright2002contextual,gilbert1990influence,zipser1996contextual}. For example, Zipser et al.~\cite{zipser1996contextual} studied the contextual modulation in the primary visual cortex of awake, behaving macaque monkeys. The work shows that the activity of a visual neuron is influenced by the stimuli placed entirely outside of its default receptive field. Furthermore, the work demonstrates that the influence of the context on the activity of a visual neuron is present even at the early stages of the visual system (V1 area).
Albright et al.~\cite{albright2002contextual} stated that for each local region of an image, the extraction of semantic meaning is only possible if information from other regions is taken into account. This clearly highlights the importance of contextual information in the natural visual systems studied in the field of neuroscience. 

Convolutional neural networks (CNNs)~\cite{lecun1989backpropagation, lecun1998gradient} represent the backbone of nearly every current computer vision task and application~\cite{chollet2017xception,he2017mask,he2016deep,he2016identity,hu2019squeeze,huang2017densely,ioffe2015batch,krizhevsky2012imagenet,lin2017focal,lin2017feature,simonyan2014very,szegedy2015going,wu2018group, xie2017aggregated,zoph2018learning}. Although the neuroscience studies mentioned above~\cite{albright2002contextual,gilbert1990influence,zipser1996contextual} clearly demonstrate the presence and the importance of contextual influence in a visual neuron of a biological visual cortex, in the current artificially-built visual systems, the core building block of CNNs, represented by the convolutional layer (with spatial filters that activate on local patterns), is not implicitly equipped with the ability of integrating contextual information. In general, the convolutional filters have a limited receptive field, usually corresponding to a $3\!\times\!3$ spatial kernel size, due to the fact that increasing the kernel size brings additional costs in terms of parameters and computational resources. There are many approaches that address the integration of contextual information, e.g.~\cite{wang2018non}, however, most of them follow the direction of integrating additional building blocks in the CNN to incorporate contextual information. However, this line of research results in additional costs for the CNN in terms of both parameters and computation, which is not in line with the neuroscience findings, pointing out that the visual system is extremely efficient and that the integration of the contextual information is an implicit capability of the visual neuron. 
Inspired by the aforementioned neuroscience studies and addressing the above limitations, this work makes the following contributions:
\vspace{-0.22cm}
\begin{itemize}
    \item We propose contextual convolution (CoConv), a direct replacement of the standard convolution that can be used at any stage in CNN architectures. CoConv is implicitly equipped with the ability of accessing contextual information at multiple levels  without increasing the demands in terms of parameters and computational cost, compared to the standard convolution (see Section~\ref{sec:coconv}).\vspace{-0.2cm}
    \item We integrate CoConv in convolutional and generative networks of various depths, presenting novel architectures based on CoConv for visual recognition and generation (see Section~\ref{sec:coresnet}).\vspace{-0.2cm}
    \item We show improved detection, recognition and generation performance obtained by CoConv over the standard convolution and competing methods on the core tasks and benchmarks for visual recognition and generation (see Section~\ref{sec:exp}).
\end{itemize}
\vspace{-0.2cm}


We underline that our approach, CoConv, is both effective and efficient. We believe that its simplicity coupled with its effectiveness generates a great potential to become widely-adopted.

\section{Related Work}
\vspace{-0.1cm}

There are numerous works with the goal of integrating contextual information in an artificial visual system. The work of Wang et al.~\cite{wang2018non} introduced a non-local block to capture contextual information in a CNN. In~\cite{hu2019squeeze}, the authors proposed a squeeze-and-excitation block to capture global information and scale each feature map accordingly. However, these works propose additional building blocks that need to be inserted in the CNN, therefore bringing significant supplementary parameters and computational costs that can negatively impact the efficiency of the visual system. In contrast, we propose to implicitly integrate the process of capturing contextual information in the core component (the convolutional layer), without increasing the number of parameters and computational costs. Furthermore, our approach can be complementary to these works, as they still need to use convolutional layers, in their overall CNN architectures.
Dilated convolution~\cite{chen2017deeplab,yu2015multi,yu2017dilated} is an approach to enlarge the receptive field of the convolution kernel. Our work makes use of dilated convolution, however, there are significant differences in the approach and usage from previous works. For example, Chen et al.~\cite{chen2017deeplab} proposed atrous spatial pyramid pooling (ASPP) to segment objects at multiple scales. There are fundamental differences that distinguish our work from that of Chen et al.~\cite{chen2017deeplab}, as explained next. First, ASPP is proposed just as a head module for image segmentation, while our approach is designed as a direct replacement of the convolution along all stages of the CNN architecture, irrespective of the visual recognition task. Second, different from ASPP, our approach is specifically designed to integrate contextual information at different levels while maintaining the same number of parameters and computational costs as the standard convolution. Importantly, as in the neuroscience findings~\cite{albright2002contextual,zipser1996contextual} showing that contextual influence is present and relevant, even in the primary visual cortex (V1 area), we integrate the contextual information within all network layers, including the early convolutional layers as well. Contrary to the neuroscience findings, Chen et al.~\cite{chen2017deeplab} employed contextual modeling only at the end of the CNN, just as an additional head before the final classification layer for semantic segmentation.
A more closely related work to our own is~\cite{yu2017dilated}, which proposes dilated residual networks by integrating dilated convolution just towards the end of the network. Thus, this work is also not in line with the neuroscience conclusion regarding the importance of the contextual modeling at the early stages of the visual cortex. Furthermore, our approach is different from that of Yu et al.~\cite{yu2017dilated}, as we employ different levels of contextual information, being able to capture information about local details and various levels of contextual information in the same time. Importantly, integrating the default approach of Yu et al.~\cite{yu2017dilated} into residual networks for image recognition drastically increases the demands in computational resources. As shown in the experiments, our approach delivers improved recognition performance without increasing the computational costs.
%

\begin{figure*}[th]
\captionsetup[subfigure]{labelformat=empty}
\centering
\subfloat[(a)]
{	
	\includegraphics[width=0.8\linewidth]{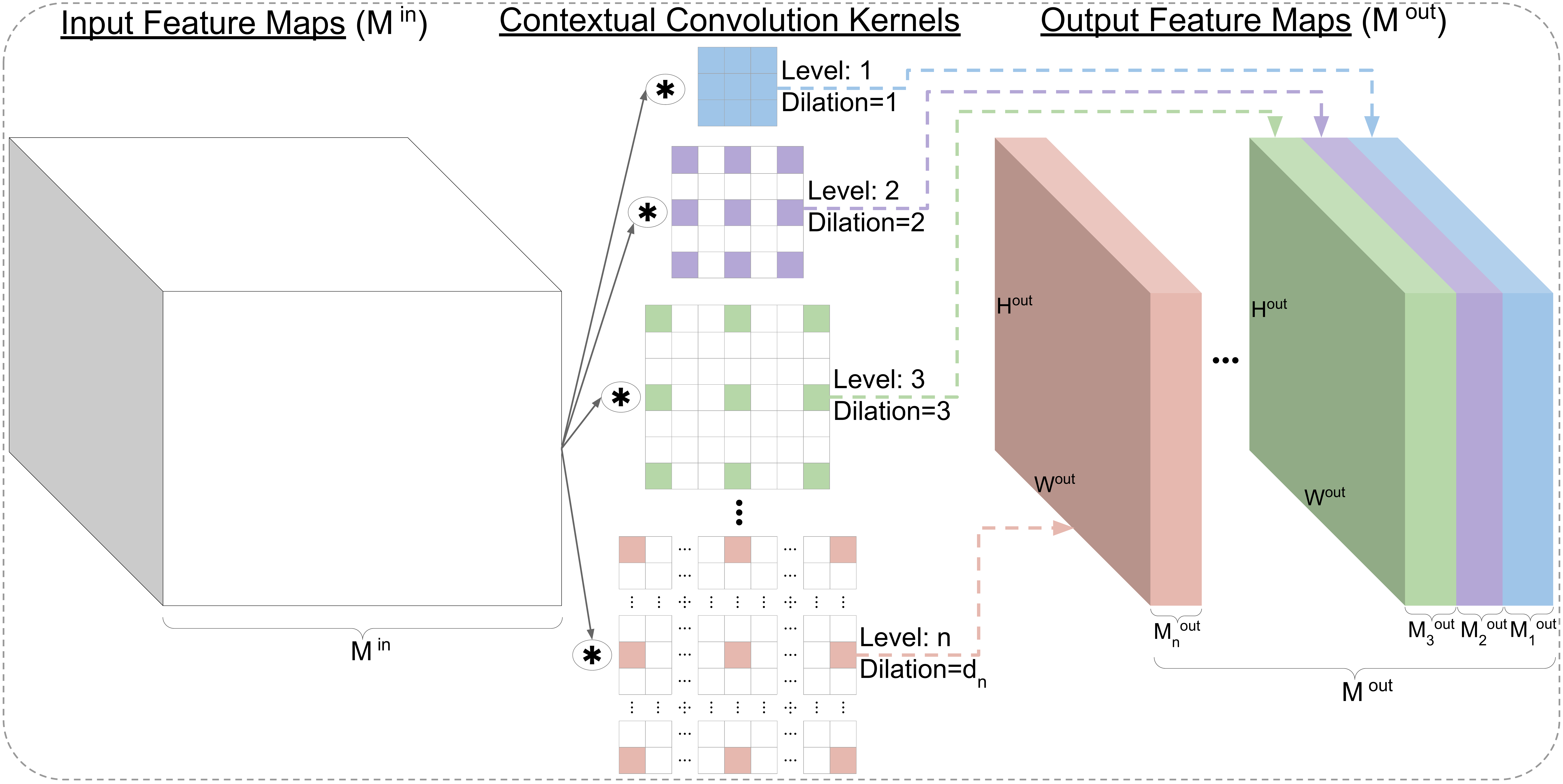}
	\label{fig:coconv}
}
\subfloat[(b)]
{
	\includegraphics[ width=0.18\linewidth]{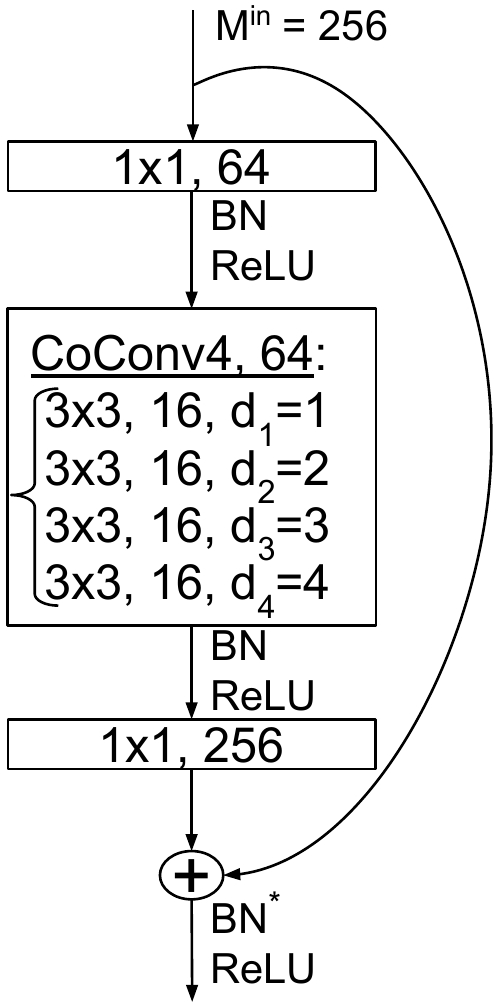}
	\label{fig:coconv_block}
}
\vspace{-0.25cm}
\caption{(a) Contextual Convolution (CoConv). Instead of using standard or dilated convolution, we propose to integrate multiple levels of kernels with different dilation ratios in the convolutional layer. At each level, we have multiple kernels. We emphasize the fact that, in this illustration, $d_1=1$, $d_2=2$ and $d_3=3$ is a coincidence that facilitates visualization, yet, in general, we do not constrain $d_i$ to be equal to $i$. Best viewed in color. (b) An example of CoConv residual building block.}
\vspace{-0.3cm}
\end{figure*}

Another contribution that is aimed at capturing context in all stages of neural architectures is represented by capsule networks (CapsNets)~\cite{sabour2017dynamic}. Although CapsNets are also backed by neurosciene studies and showed promising results on small data sets such as MNIST and CIFAR-10, their low accuracy gains come with a large computational cost. Furthermore, since their introduction by Sabour et al.~\cite{sabour2017dynamic}, CapsNets have failed to show their effectiveness on very deep neural networks and on large image recognition benchmarks, mostly due to their extremely large computational costs. Different from~\cite{sabour2017dynamic}, we present empirical evidence showing that CoConv improves the accuracy of very deep models, e.g.~ResNet-152, on very large benchmarks, e.g.~ImageNet, at no additional cost. Another plus is that our contribution is fairly easy to implement, having the right ingredients (simple, effective, no additional computational cost) to be widely adopted by the community.

\section{Contextual Convolution}
\label{sec:coconv}
\vspace{-0.1cm}

The standard convolution in state-of-the-art CNN architectures uses a single type of kernel with a fixed receptive field, usually corresponding to a kernel size of $3\!\times\!3$, since increasing the kernel size brings additional costs in terms of the number of learnable parameters and computational time, respectively. The number of learnable parameters (weights) and FLOPs (floating point operations) for the standard convolution can be computed as:
\begin{equation}
 \label{eq:conv}
\begin{array}{l}
    \mbox{params} = M^{in} \cdot K^w \cdot K^h \cdot M^{out}, \\
    \mbox{FLOPs} =  M^{in} \cdot K^h \cdot K^w \cdot M^{out} \cdot  W^{out} \cdot H^{out},
\end{array}
\end{equation}
where, $M^{in}$ and $M^{out}$ represent the number of input and output feature maps, $K^w$ and $K^h$ are the width and height of the convolution kernel, and finally, $W^{out}$ and $H^{out}$ are the width and height of the output feature maps. For the sake of simplicity, we ignored the bias terms and hyperparameters such as stride and padding
in Equation~(\ref{eq:conv}).

Contextual convolution (CoConv), illustrated in Figure~\ref{fig:coconv},  receives a number of input feature maps $M^{in}$, over which we apply different levels $L=\{1,2,3,...,n\}$ of convolution kernels with varying dilation ratios $D=\{d_1, d_2, d_3, ..., d_n\}$. In other words, the kernels at level $i$ have the dilation ratio $d_i$, $\forall i \in L$.
By gradually increasing the dilation ratio (basically introducing increasingly larger ``holes'' into the kernels), the filters can have access to increasingly broader contextual information. As we increase the dilation ratio, the kernels become sparser, thus, being applied over the input feature maps in a sparse pattern, skipping elements in the computation. As depicted in Figure~\ref{fig:coconv}, only the colored spatial locations of the kernels are involved in the computation of the output feature maps. Thus, each level of dilated kernels maintains a similar number of parameters and FLOPs, while increasing the spatial receptive field to integrate more contextual information. The kernels with lower dilation ratios are responsible for capturing information about local details from the input feature maps, while the kernels with higher dilation ratios are empowered with the ability of incorporating contextual information for helping the recognition process.   
At each level $i$, the kernels provide a number of output feature maps $M_i^{out}$, for all $i \in L$,
each map having the width $W^{out}$ and the height $H^{out}$. Hence, the total number of learnable parameters and FLOPs of CoConv is computed as follows:
\begin{equation}
 \label{eq:coconv}
 \hspace{-0.4cm}
\begin{split}
    \mbox{params}\!&\!=\!M^{in}\!\cdot\!(K^w\!\cdot\!K^h)^{(d_1)}\!\cdot \!M_1^{out} + \\
    &... + M^{in}\!\cdot\!(K^w\!\cdot\!K^h)^{(d_n)}\!\cdot\! M_n^{out},\\
    \mbox{FLOPs}\!&=\!M^{in}\!\cdot\!(K^w\!\cdot\!K^h)^{(d_1)}\!\cdot\!M_1^{out}\!\cdot\!W^{out}\!\cdot\!H^{out}\!+ \\ &...\!+\!M^{in}\!\cdot\!(K^w\!\cdot\!K^h)^{(d_n)}\!\cdot\!M_n^{out}\!\cdot\!W^{out}\!\cdot\!H^{out},
\end{split}
\end{equation}
where, $(K^w\cdot K^h)^{(d_i)}$ refers to the spatial size of the kernel of width $w$ and height $h$ (basically, how many spatial locations of the kernel are involved in the computation of an output feature map), $d_i$ points to the dilation ratio used for the kernels at level $i$, and:
\begin{equation}\label{eq:num_kers}
M^{out} = \sum_{i=1}^n M_i^{out}, \forall i \in L.
\end{equation}

Although we use multiple levels of kernels with different dilation ratios, the total number of kernels in CoConv is equal to the total number of kernels in the standard convolution, as it results from Equation~(\ref{eq:num_kers}). We emphasize that all levels and kernels in CoConv are independent, allowing parallel execution, just as in a standard convolutional layer.
CoConv is a direct replacement of the standard convolution with the capability of integrating contextual information.
Moreover, as formally presented in Equations~(\ref{eq:coconv}), the number of learnable parameters and FLOPs involved in CoConv is equal to those involved in the standard convolution from Equation~(\ref{eq:conv}). We emphasize that the number of dilation levels of CoConv can be adjusted, for instance, based on the resolution of the feature maps. We present various practical examples in the next section.


\section{Contextual Convolutional Neural Networks}
\label{sec:coresnet}


We describe some neural architectures based on contextual convolution (CoConv). First, we integrate CoConv in the widely-used residual networks (ResNets)~\cite{he2016deep}. ResNets can be split into four main stages depending on the proximity of the layers with respect to the input and, implicitly, on the resolution of the feature maps, as shown in Table~\ref{table:net}. 
Figure~\ref{fig:coconv_block} shows an example of a CoConv residual building block used in the first stage of a network. 
The CoConv residual block uses a $1\!\times\!1$ convolution to reduce the number of feature maps to $64$, followed by a CoConv with four levels to capture contextual information. Each CoConv level has a different dilation ratio. The number of output feature maps at each level is $16$, regardless of the dilation rate. Then, a $1\!\times\!1$ convolution is used to restore the number of feature maps to $256$. As in the standard residual blocks, batch normalization (BN)~\cite{ioffe2015batch} and Rectified Linear Unit (ReLU) activations~\cite{nair2010rectified} follow each convolutional block.

\begin{table}[t]
\renewcommand{\arraystretch}{1.2}
\addtolength{\tabcolsep}{-3.5pt}
\centering 
\begin{tabular}{c|c|c|c}
\toprule
{stage}                  & 
output                 & 
ResNet-50             & 
CoResNet-50  \\ 
\hline

\multirow{2}{*}{}                 &
112$\times$112                & 
7$\times$7, 64, $s\!=\!2$     & 
7$\times$7, 64, $s\!=\!2$  \\
\cline{2-4}
 & 
 56$\times$56& 
3$\times$3 maxpool& 3$\times$3 maxpool\\  
& & $s\!=\!2$& $s\!=\!2$  \\  
 \hline
 
1&
56$\times$56 & 
\hspace{-0.1cm}\bsplitcell{1$\times$1, 64\\ 3$\times$3, 64\\ 1$\times$1, 256}$\!\times$3 &
\hspace{-0.1cm}\bsplitcell{1$\times$1, 64\\ \underline{CoConv4, 64:}\\  \hspace{-0.1cm}\bsplitcell{3$\times$3, 16, $d_1$=1\\ 3$\times$3, 16, $d_2$=2 \\ 3$\times$3, 16, $d_3$=3 \\ 3$\times$3, 16, $d_4$=4}\hspace{-0.1cm} \\ 1$\times$1, 256}$\!\times$3 \\ 
\hline

2 &
28$\times$28 &
\hspace{-0.1cm}\bsplitcell{1$\times$1, 128\\ 3$\times$3, 128\\ 1$\times$1, 512}$\!\times$4 &
\hspace{-0.1cm}\bsplitcell{1$\times$1, 128\\ \underline{CoConv3, 128:}\\ \hspace{-0.1cm}\bsplitcell{3$\times$3, 64, $d_1$=1 \\ 3$\times$3, 32, $d_2$=2 \\ 3$\times$3, 32, $d_3$=3}\hspace{-0.1cm} \\ 1$\times$1, 512}$\!\times$4  \\ 
\hline

3 &
14$\times$14 &
\hspace{-0.1cm}\bsplitcell{1$\times$1, 256\\ 3$\times$3, 256\\ 1$\times$1, 1024}$\!\times$6 &
\hspace{-0.1cm}\bsplitcell{1$\times$1, 256\\ \underline{CoConv2, 256:}\\  \hspace{-0.1cm}\bsplitcell{3$\times$3, 128, $d_1$=1 \\ 3$\times$3, 128, $d_2$=2}\hspace{-0.1cm} \\ 1$\times$1, 1024}$\!\times$6  \\ \hline

4 &
7$\times$7 &
\hspace{-0.1cm}\bsplitcell{1$\times$1, 512\\ 3$\times$3, 512\\ 1$\times$1, 2048}$\!\times$3 &
\hspace{-0.1cm}\bsplitcell{1$\times$1, 512\\ \underline{CoConv1, 512:}\\  \hspace{-0.1cm}\bsplitcell{3$\times$3, 512, $d_1$=1}\hspace{-0.1cm} \\ 1$\times$1, 2048}$\!\times$3 \\ 
\hline

 &
1$\times$1 &
\begin{tabular}[c]{@{}c@{}}global avgpool\\ 1000-d fc\end{tabular} &
\begin{tabular}[c]{@{}c@{}}global avgpool\\ 1000-d fc\end{tabular} \\ 
\hline

\multicolumn{2}{c|}{\# params}                  &
{25.56}  $\times$ $10^6$                 &
{25.56}  $\times$ $10^6$                   \\
\hline

\multicolumn{2}{c|}{FLOPs}                      & 
{4.14}  $\times$ $10^9$                   &
{4.14}  $\times$ $10^9$  \\
\bottomrule
\end{tabular}
\vspace{-0.2cm}
\caption{\small A side-by-side comparison of ResNet-50 and CoResNet-50. Although we illustrate our updates on ResNet-50, the changes can be analogously operated on models of different depths, e.g.~ResNet-152.}
\label{table:net}
\vspace{-0.5cm}
\end{table}

\begin{table}[t]
\renewcommand{\arraystretch}{1.2}
\addtolength{\tabcolsep}{-2.0pt}
\centering 
\begin{tabular}{c|c|c}
\toprule 
output                 & 
ProGAN                 & 
CoProGAN   \\ 
\hline
 
4$\times$4                        & 
3$\times$3, 512               &  
3$\times$3, 512              \\  
 
\hline
8$\times$8              & 
2$\times$2 upsample              &  
2$\times$2 upsample             \\  
\hline

8$\times$8              & 
3$\times$3, 512    &  
\hspace{-0.1cm}\bsplitcell{3$\times$3, 256, $d_1$=1\\ 3$\times$3, 256, $d_2$=2} \\  

\hline
16$\times$16              & 
2$\times$2 upsample              &  
2$\times$2 upsample            \\  
\hline

16$\times$16              & 
3$\times$3, 512    &  
\hspace{-0.1cm}\bsplitcell{3$\times$3, 172, $d_1$=1\\ 3$\times$3, 170, $d_2$=2 \\ 3$\times$3, 170, $d_3$=3} \\  

\hline
32$\times$32             & 
2$\times$2 upsample              &  
2$\times$2 upsample             \\  

\hline
32$\times$32              & 
3$\times$3, 512    &  
\hspace{-0.1cm}\bsplitcell{3$\times$3, 128, $d_1$=1\\ 3$\times$3, 128, $d_2$=2 \\ 3$\times$3, 128, $d_3$=3 \\ 3$\times$3, 128, $d_4$=4} \\  
  
\hline
64$\times$64             & 
2$\times$2 upsample              &  
2$\times$2 upsample             \\    

\hline
64$\times$64              & 
3$\times$3, 256    &  
\hspace{-0.1cm}\bsplitcell{3$\times$3, 64, $d_1$=1\\ 3$\times$3, 64, $d_2$=2 \\ 3$\times$3, 64, $d_3$=3 \\ 3$\times$3, 64, $d_4$=4} \\  

\hline
128$\times$128             & 
2$\times$2 upsample              &  
2$\times$2 upsample             \\   

\hline
128$\times$128               & 
3$\times$3, 128    &  
\hspace{-0.1cm}\bsplitcell{3$\times$3, 32, $d_1$=1\\ 3$\times$3, 32, $d_2$=2 \\ 3$\times$3, 32, $d_3$=3 \\ 3$\times$3, 32, $d_4$=4} \\  
\hline
128$\times$128             & 
1$\times$1, 3               &  
1$\times$1, 3             \\  
\hline
{\# params}                  &
27.21  $\times$ $10^6$                 &
27.21  $\times$ $10^6$                   \\
\hline

{FLOPs}                      & 
54.76  $\times$ $10^9$                    &
54.76  $\times$ $10^9$   \\
\bottomrule
\end{tabular}
\vspace{-0.2cm}
\caption{\small A side-by-side comparison of ProGAN and CoProGAN.}
\label{table:gan}
\vspace{-0.5cm}
\end{table}

Our network for image classification, termed contextual residual network (CoResNet), is formally presented in Table~\ref{table:net}. Although we illustrate our updates on ResNet-50, thus obtaining CoResNet-50, the changes can be analogously operated on models of different depths. Since the size of the feature maps decreases as the layers are farther away from the input, we also adapt the number of levels in our CoConv layers with respect to the resolution of the feature maps. Hence, the first main stage of the network uses four levels in the CoConv layers, with different dilation ratios. Further, the second stage uses three levels in the CoConv layers, while the third stage uses two levels. As the spatial resolution of the feature maps in the last stage is reduced to $7\!\times\!7$, we consider that using multiple dilation ratios is no longer justified. Thus, the last stage employs just one level of CoConv. In the experimental section, we provide an ablation study on the number of levels in the CoConv layers. However, the number of levels can be tuned for each particular task or application, based, for instance, on the resolution of the feature maps along the network. Based on empirical evidence, we consider that our network has improved recognition capabilities compared to the standard ResNet, as the convolution is equipped with the ability of integrating contextual information at multiple levels and, as can be seen in Table~\ref{table:net}, CoConv does not add any additional weights nor it enlarges the computational cost compared to the original network. 

We also show the benefits of integrating CoConv in a generative adversarial network (GAN) \cite{Goodfellow-NIPS-2014}. We specifically consider progressive GAN (ProGAN) \cite{karras-ICLR-2018}, a model that generates high-resolution images starting with a low-resolution output ($4\!\times\!4$ pixels) and gradually adding layers to the network to produce a high-resolution output (up to $1024\!\times\!1024$ pixels). We used the exact same architecture described in \cite{karras-ICLR-2018}, only replacing the convolutional layers with CoConv layers. The ProGAN and CoProGAN generator for the CelebA data set \cite{Liu-ICCV-2015} is illustrated in Table~\ref{table:gan}. The CoProGAN generator designed for the final output of $128\!\times\!128$ pixels starts with a feature map size of $4\!\times\!4$ pixels. Hence, at the first layer, we only use a dilation rate of $1$. When the output size increases to $8\!\times\!8$ pixels, we add two dilation rates of $1$ and $2$. Similarly, we add three dilation rates ($1$, $2$ and $3$), when the output size is $16\!\times\!16$ pixels. When the output size is $32\!\times\!32$, $64\!\times\!64$ or $128\!\times\!128$ pixels, we use four dilation rates of  $1$, $2$, $3$ and $4$, regardless of the size of the output. The number of filters in each CoConv layer matches the number of filters in the corresponding conv layer from ProGAN. Thus, the number of parameters and FLOPs in ProGAN and CoProGAN are identical.

\vspace{-0.1cm}
\section{Experiments \label{sec:exp}}
\vspace{-0.12cm}
\subsection{Experimental setup}  
\vspace{-0.08cm}

We perform object recognition experiments on the ImageNet Large Scale Visual Recognition Challenge (ILSVRC)~\cite{russakovsky2015imagenet}, which is one of the most popular benchmarks in visual recognition. The ImageNet data set consists of 1000 classes of objects, 1.28 million training images and 50K validation images. As common, we report both the top-1 and top-5 error rates. We follow the standard settings in \cite{goyal2017accurate, he2016deep, he2016identity} and employ the Stochastic Gradient Descent (SGD) optimizer with a standard momentum rate of $0.9$ and a weight decay of $0.0001$. We perform the training for $90$ epochs, starting with a learning rate of $0.1$, reducing it by $1/10$ at the 30-th, 60-th and 80-th epochs, similarly to \cite{goyal2017accurate,he2016deep}. Each model is trained using 8 GPUs. We use the standard mini-batch size of $256$ for training and data augmentation as in~\cite{goyal2017accurate,szegedy2015going}, training and testing on $224$$\times$$224$ image crops.

For the object detection task, we consider the MS COCO data set~\cite{lin2014microsoft}, which contains 80 object categories. We use COCO 2017 train (118K images) for training and COCO 2017 val (5K images) for testing. We train each model for $130$ epochs on 8 GPUs using mini-batches of $32$ examples, resulting in 60K training iterations. The training is performed using the SGD optimizer with momentum $0.9$, weight decay $0.0005$, with the learning rate $0.02$ (reduced by $1/10$ before the 86-th and 108-th epochs). We also use a linear warm-up in the first epoch, following~\cite{goyal2017accurate}. For data augmentation, we perform random crop as in~\cite{liu2016ssd}, color jittering and horizontal flip. We consider an input image size of $300\!\times\!300$ pixels. As evaluation metrics, we report the average precision (AP) and the AP@IoU=0.5.

We conduct image generation experiments on CIFAR-10 \cite{Krizhevsky-TR-2009} and CelebA \cite{Liu-ICCV-2015}, considering only fully unsupervised (not class conditional) GAN models. The CIFAR-10 training set is composed of 50K images of $32\!\times\!32$ pixels, while the CelebA training set is formed of 202K images of $128\!\times\!128$ pixels. Following Karras et al.~\cite{karras-ICLR-2018}, the optimization is performed using the Adam \cite{Kingma-ICLR-1015} optimizer with $\beta_1=0$, $\beta_2=0.99$, $\epsilon=10^{-8}$ and the learning rate set to $10 ^ {-3}$. We train each model until the discriminator sees $12$ million real images in total. Each model is trained on a single GeForce GTX 3090 GPU. As evaluation measures, we report the Inception Score (IS) \cite{Salimans-NIPS-2016} and the Fr\'{e}chet Inception Distance (FID)~\cite{Heusel-NIPS-2017} for CIFAR-10, and the multiscale structural similarity index measure (MS-SSIM) \cite{Wang-ACSSC-2003} and the Sliced Wasserstein Distance (SWD) \cite{Rabin-SSVMCV-2012} for CelebA. To reproduce the results of ProGAN, we used the official TensorFlow implementation available at {\small \url{https://github.com/tkarras/progressive\_growing\_of\_gans}}.

\vspace{-0.06cm}
\subsection{Ablation experiments on CoConv dilation levels}
\vspace{-0.04cm}

\begin{table}[t]
  \centering
  \begin{tabular}{lcccc}
    \toprule
   CoConv levels & top-1     & top-5 & params & GFLOPs\\
    \midrule
    (1,1,1,1) &  23.88 & 7.06 & 25.56 & 4.14   \\
    (2,2,2,1) &  23.24 & 6.79 & 25.56 & 4.14   \\
    (3,3,2,1) &  23.23 & 6.66 & 25.56 & 4.14   \\
    (3,3,3,3) &  23.33 & 6.71 & 25.56 & 4.14   \\
    (4,3,2,1)&  22.73 & 6.49 & 25.56 & 4.14   \\
    top(4,3,2,1) &  25.58 & 8.05 & 25.56 & 4.14   \\
    \bottomrule
  \end{tabular}
  \vspace{-0.2cm}
  \caption{Ablation experiments on ImageNet with various CoResNet-50 configurations, considering different numbers of dilation levels in the CoConv residual blocks. The configuration $(1,1,1,1)$ corresponds to ResNet-50~\cite{he2016deep}. Lower values are better.}
  \label{table:ablation}
  \vspace{-0.2cm}
\end{table}

In Table~\ref{table:ablation}, we present ablation experiments with different configurations generated by varying the number of dilation levels in the ConConv residual blocks corresponding to each stage of the network. The first column indicates the number of levels in the CoConv residual block used in each of the four main stages of the network. The configuration $(1,1,1,1)$ refers to the case where a single CoConv level with dilation $d_1\!\!=\!\!1$ is used in all four stages, thus being completely equivalent to the original ResNet~\cite{he2016deep}, which is the baseline in our object recognition experiments. 

Integrating our CoConv with two levels for the first three stages of the network, resulting in the configuration $(2,2,2,1)$, significantly improves the top-1 error rate from $23.88\%$ to $23.24\%$, while maintaining the same number of parameters and FLOPs. In general, adding more levels to our CoConv residual blocks further improves the results. 

We obtain the best results with the configuration $(4,3,2,1)$ for the number of levels in the CoConv blocks involved in the four main stages of the network. We hereby note that we performed experiments with even more levels of CoConv, but we did not notice further significant improvements in terms recognition performance. Hence, we find the configuration $(4,3,2,1)$ optimal for an input size of $224\!\times\!224$ pixels. If the input size would be higher, then perhaps another configuration considering more dilation levels can provide even further improvements. All in all, the flexibility of adapting the CoConv levels for each network stage with respect to the input resolution is an important strong point of our approach.

\begin{table}[t]
\addtolength{\tabcolsep}{-4.0pt}
  \centering
  \begin{tabular}{lccccc}
    \toprule
   Network & stride & top-1     & top-5 & params & GFLOPs \\
    \midrule
    ResNet-50~\cite{he2016deep} & 32 &  23.88 & 7.06 & 25.56 & 4.14    \\
    NL ResNet-50~\cite{wang2018non} & 32 & 22.91 & 6.42 & 36.72 & 6.18 \\
    CoResNet-50~[ours] & 32 &  22.73 & 6.49 & 25.56 & 4.14 \\
    \midrule
    DRN-50~\cite{yu2017dilated} & 8 & 22.44 & 6.47 & 25.56 & 19.20 \\
    CoResNet-50~[ours] & 8 &  21.93 & 6.17 & 25.56 & 19.20 \\
    \bottomrule
  \end{tabular}
  \vspace{-0.2cm}
  \caption{Comparison of CoResNet with closely related works~\cite{wang2018non,yu2017dilated} on ImageNet, considering neural models of 50 layers in all cases. Lower values are better.}
  \label{table:com_rel_work}
\vspace{-0.35cm}
\end{table}

\begin{figure*}[t]
\centering
\begin{tabular}{cccc}
\subfloat{\includegraphics[width=0.31\linewidth]{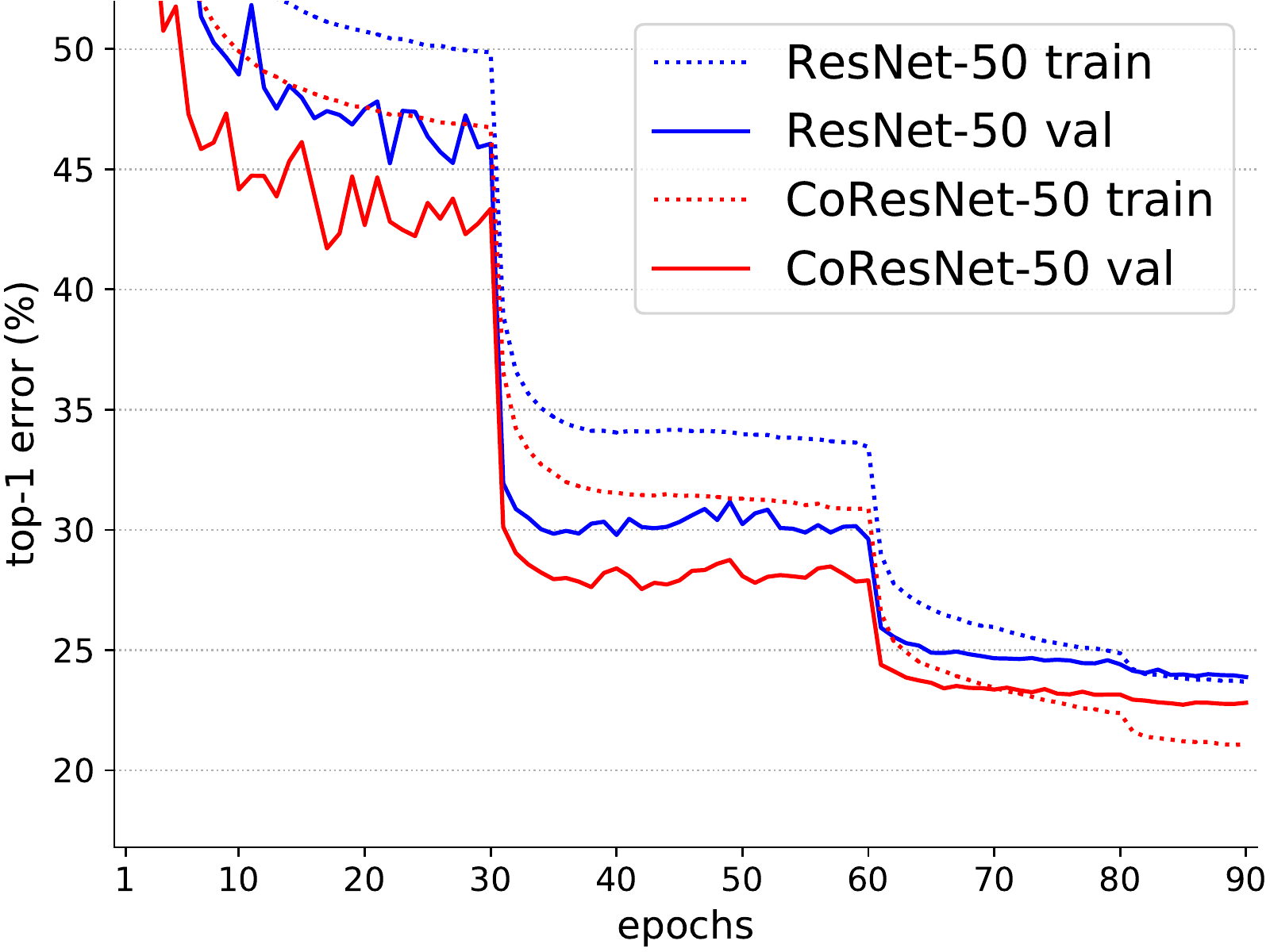}} &
\subfloat{\includegraphics[width=0.31\linewidth]{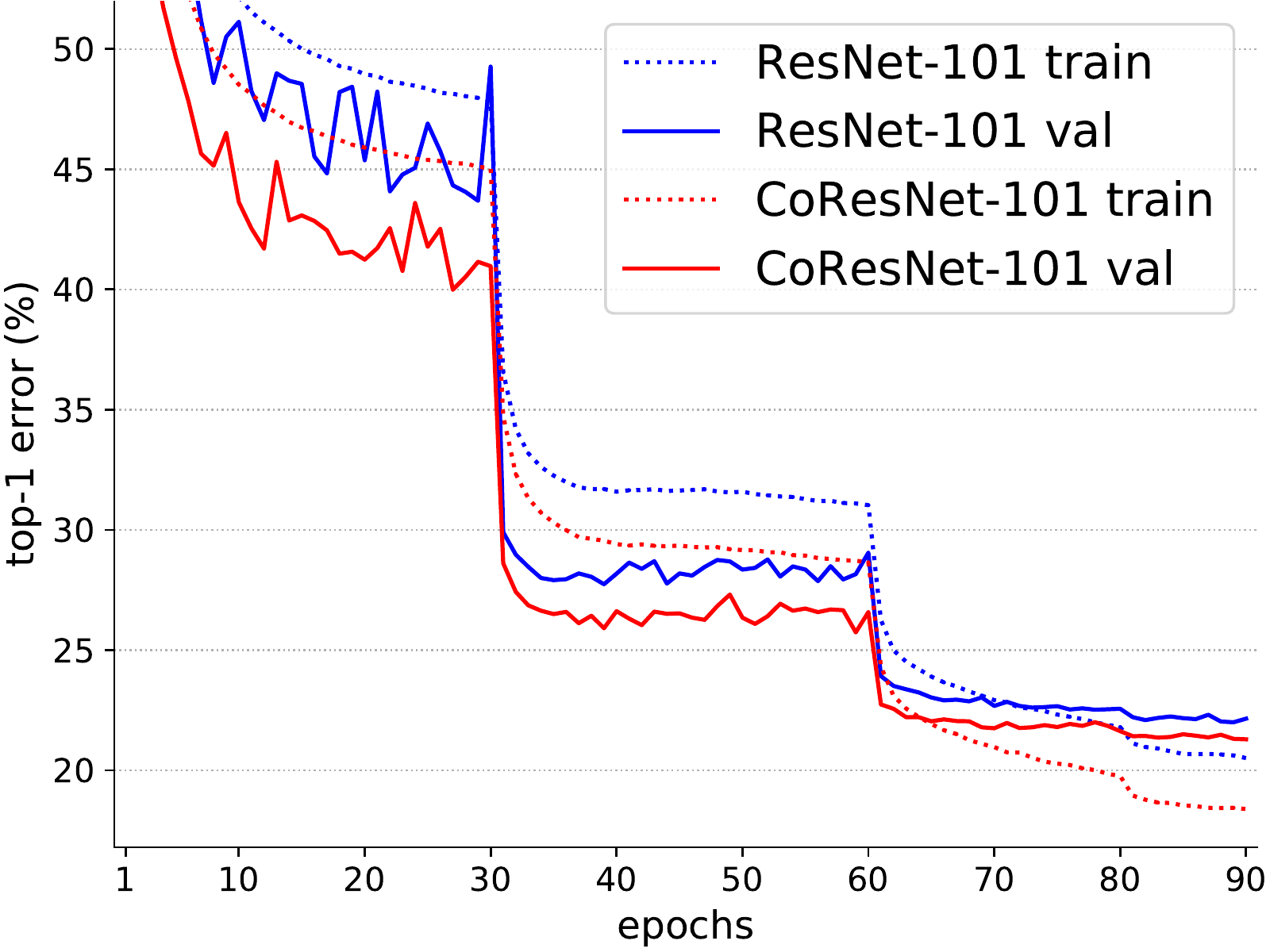}} &
\subfloat{\includegraphics[width=0.31\linewidth]{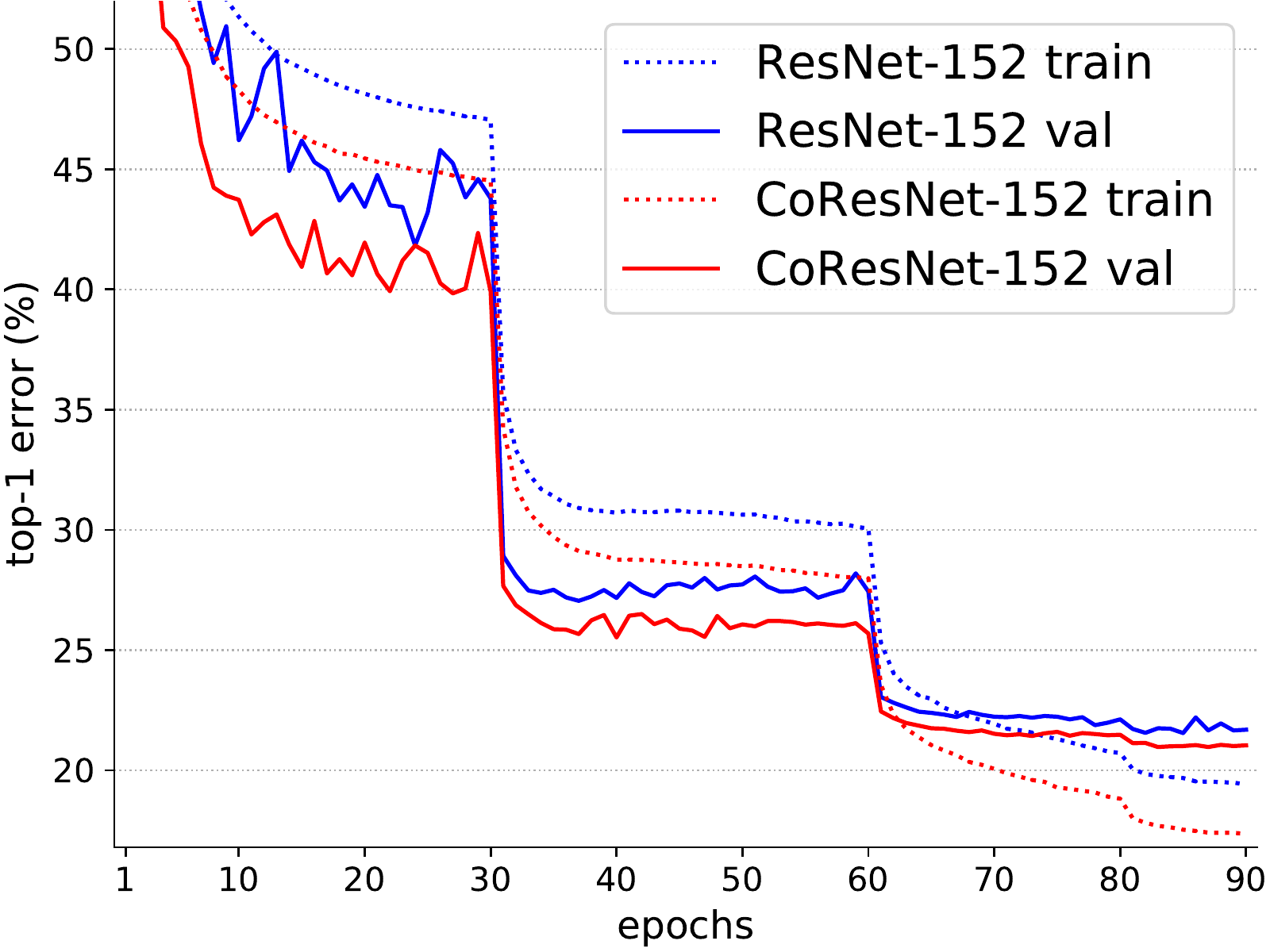}}
\end{tabular}
\vspace{-0.45cm}
\caption{Training and validation curves on ImageNet for ResNet and CoResNet architectures of 50, 101 and 152 layers, respectively. Best viewed in color.}
\label{fig:curves_coresnet}
\vspace{-0.1cm}
\end{figure*}

\addtolength{\tabcolsep}{-1.2pt}
\begin{table*}[t]
\centering
\begin{tabular}{l cccc cccc cccc}
\toprule
\multirow{2}{*}{Network}&\multicolumn{4}{c}{network depth: 50 }&\multicolumn{4}{c}{network depth: 101 }&\multicolumn{4}{c}{network depth: 152}\\  \cmidrule(r){2-5} \cmidrule(r){6-9} \cmidrule(r){10-13}
&top-1&top-5&\scalebox{0.8}{params}&\scalebox{0.7}{GFLOPs}&top-1&top-5&\scalebox{0.8}{params} &\scalebox{0.7}{GFLOPs}&top-1&top-5&\scalebox{0.8}{params} &\scalebox{0.7}{GFLOPs}\\
\midrule
ResNet~\cite{he2016deep}&23.88&7.06&25.56&4.14&22.00&6.10&44.55&7.88 &21.55&5.74&60.19&11.62\\ 
pre-act. ResNet \cite{he2016identity}&23.77&7.04&25.56&4.14&22.11&6.26&44.55&7.88& 21.41&5.78&60.19&11.62 \\
SE ResNet \cite{hu2019squeeze} & 22.74 & 6.37 & 28.07 & 4.15 & 21.31 & 5.79 & 49.29 & 7.90 & 21.38 & 5.80& 66.77 & 11.65 \\
NL ResNet~\cite{wang2018non}  & 22.91 & 6.42 & 36.72 & 6.18 & 21.40 & 5.83 & 55.71& 9.91 & 21.91 & 6.11 & 71.35& 13.66 \\
 {\bf CoResNet} [ours] &  22.73 & 6.49 & 25.56 & 4.14 & 21.29&5.72&44.55&7.88 &20.97&5.48&60.19&11.62\\ 
\bottomrule
\end{tabular}
\vspace{-0.2cm}
\caption{ImageNet results of CoResNet in comparison with other state-of-the-art methods \cite{he2016identity,hu2019squeeze,wang2018non}, considering architectures on different depths, ranging from 50 layers to 152 layers.}
\label{table:comp_depth}
\vspace{-0.3cm}
\end{table*}
\addtolength{\tabcolsep}{1.2pt}

To show the importance of having different levels of dilation in CoConv, we include the configuration top$(4,3,2,1)$ in Table~\ref{table:ablation}, which considers only the highest level of dilation in each stage of the network. For instance, in the first stage, only the convolution with dilation ratio 4 is used, the second stage uses only the convolution with dilation 3, and so on. For a fair comparison, we stress out that the number of filters in each CoConv layer is always equal to the number of filters in the original ResNet model. Regarding the configuration top$(4,3,2,1)$, we can notice a significant drop in recognition performance. This is basically the opposite case of the baseline $(1,1,1,1)$. We can observe from the results that both (extreme) cases have significantly lower recognition performance than CoConv with multiple levels. This is due to the fact that the baseline $(1,1,1,1)$ is only able to capture information about local details (as it uses the lowest dilation), without being equipped with the ability to capture contextual information. On the other side, the case top$(4,3,2,1)$ is only able to capture contextual information, lacking the ability of capturing information about local details. This set of experiments proves an important and strong point of our approach: CoConv captures information regarding both local details and global context, providing a more complete visual representation which improves the recognition performance.

\subsection{Comparison of CoResNet to closely related works}

In Table~\ref{table:com_rel_work}, we present the results of closely related methods \cite{wang2018non,yu2017dilated} by training all neural networks with the same standard settings for providing a direct and fair comparison. All methods are applied on top of the 50-layer deep residual network. First, we observe that our CoResNet-50 outperforms more complex architectures, such as non-local (NL) networks~\cite{wang2018non}.
It is important to highlight the increase in the number of learnable parameters and computational costs brought by the introduction of the non-local block, while our CoResNet-50 maintains the same number of parameters and computational cost as the baseline ResNet-50~\cite{he2016deep}.

The dilated residual network (DRN) proposed by Yu et al.~\cite{yu2017dilated} requires to decrease the overall stride of the network from the default 32 to 8. Basically, DRN does not use downsampling of the feature maps for the last two stages of the network. In Table~\ref{table:com_rel_work}, we can observe that DRN has a significant impact in increasing the requirements in terms of computational resources, increasing the GFLOPs from $4.14$ to $19.20$. This significant increase in computational cost makes DRN not feasible for standard image classification as, for instance, increasing the depth of the baseline ResNet~\cite{he2016deep} from 50 to 101 layers provides a top-1 error improvement from $23.88\%$ to $22.00\%$, while the requirements in GFLOPs increase from $4.14$ to only $7.88$. In the same time, DRN-50 improves the baseline top-1 error from $23.88\%$ to $22.44\%$ at a higher computational cost. 
In comparison, our CoResNet-50 can improve the recognition performance of the baseline without impacting the demands in terms of computational resources. 

To make a direct comparison between our approach and DRN~\cite{yu2017dilated} under the same number of parameters and FLOPs, we also perform an experiment by setting the stride of CoResNet-50 to 8 instead of 32.  As shown in Table~\ref{table:com_rel_work}, our approach provides improved recognition performance in comparison to DRN. As another evidence that our approach is superior to DRN, we can link the DRN results from Table~\ref{table:com_rel_work} with our configuration $top(4,3,2,1)$ from Table~\ref{table:ablation}. More precisely, DRN and $top(4,3,2,1)$ are from the same category of methods, as both use only the top dilation for convolution. We have already shown that this is not the optimal case for attaining good recognition performance, as it is necessary to have kernels that can capture detailed (local) information, as well as kernels that capture contextual information. We conjecture that the range of kernels from local to contextual is important for visual perception, as different levels of kernels bring complementary information into the visual system. 

\begin{table}[t]
\addtolength{\tabcolsep}{-3pt}
  \centering
  \begin{tabular}{lcccc}
    \toprule
   Backbone& AP     & AP@IoU=0.5 & params & GFLOPs\\
    \midrule
    ResNet-50 &  26.20 & 43.97 & 22.89 & 20.92   \\
    CoResNet-50 &  28.63 & 46.71 & 22.89 & 20.92  \\
    \midrule
    ResNet-101 &  29.58& 47.69 & 41.89 & 48.45  \\
    CoResNet-101 &  31.19 & 49.89 & 41.89 & 48.45   \\

    \bottomrule
  \end{tabular}
  \vspace{-0.2cm}
  \caption{Results of SSD with various ResNet and CoResNet backbones of 50 or 101 layers on the MS COCO data set, for input images of $300\times300$ pixels. Higher AP and AP@IoU=0.5 values are better.}
  \label{table:coco}
\vspace{-0.4cm}
\end{table}

\begin{table*}[t]
  \centering
  \begin{tabular}{lcccccccc}
    \toprule
               &  \multicolumn{2}{c}{CIFAR-10} & \multicolumn{5}{c}{CelebA}\\
    \cmidrule(r){2-3} \cmidrule(r){4-9}
    Method     &   IS   & FID & MS-SSIM & \multicolumn{5}{c}{SWD $\times 10^3$}\\ 
    \cmidrule(r){5-9}
               &        &     &         &     128  &  64 & 32 & 16 & Avg. \\
    \midrule
    ProGAN \cite{karras-ICLR-2018}    &  7.60$\pm$0.09 & 20.70 & 0.2894 &   3.65  &   2.48  &   2.66  &   7.29  &  4.02\\  
    CoProGAN   &  7.71$\pm$0.06 & 19.66 & 0.2875 &   3.29  &   2.43  &   2.27  &   5.35  &  3.34\\  
    \bottomrule
  \end{tabular}
  \vspace{-0.2cm}
  \caption{ProGAN versus CoProGAN results on CIFAR-10 and CelebA. Higher IS values are better. Lower FID, MS-SSIM and SWD are better.}
  \label{table:image_generation}
\end{table*}

\begin{figure*}[!t]
  \centering
    \includegraphics[width=1.00\linewidth]{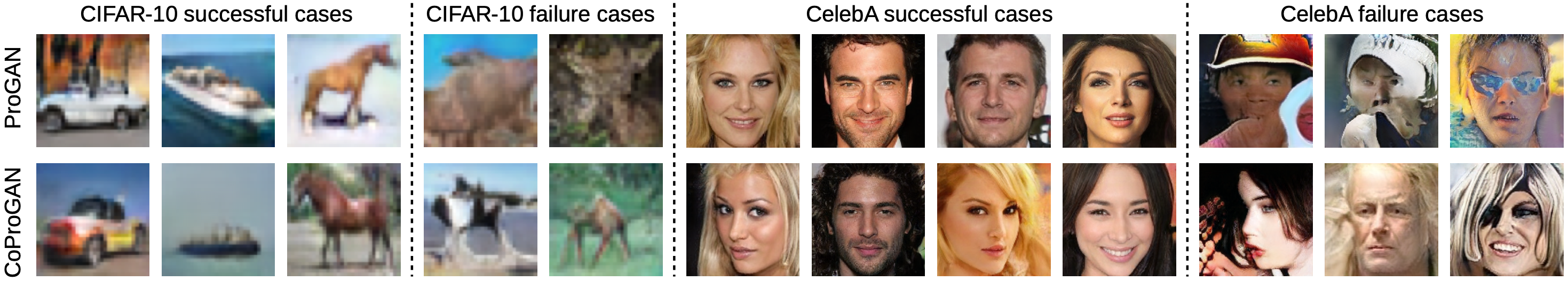}
\vspace{-0.5cm}
  \caption{Examples generated by ProGAN and CoProGAN, selected by a human annotator from a set of 50 images generated from CIFAR-10 and CelebA, respectively. Best viewed in color.\label{fig:image_generation}}
  \vspace{-0.3cm}
\end{figure*}

\subsection{Comparison of CoResNet on other architectures of different depths}

In Figure~\ref{fig:curves_coresnet}, we present the training and validation learning curves on ImageNet, considering ResNet and CoResNet architectures of 50, 101 and 152 layers, respectively. Comparing our CoResNet with the baseline ResNet~\cite{he2016deep}, we can see an improved convergence during training, this being due to the fact that CoConv provides a more complete visual representation of the input. 

In Table~\ref{table:comp_depth}, we provide the comparative results between CoResNet and several other works~\cite{he2016deep,he2016identity,hu2019squeeze,wang2018non}, considering neural networks of 50, 101 and 152 layers deep, respectively. CoResNet outperforms the baseline ResNet~\cite{he2016deep} on all network depths. We also outperform the pre-activation ResNet~\cite{he2016identity} by a consistent margin, while maintaining the same number of parameters and computational cost. 
In terms of the top-1 error rate, we outperform the non-local block of Wang et al.~\cite{wang2018non} on all three tested network depths, namely 50, 101 and 152. We qualify our results as even more impressive, considering that the work of Wang et al.~\cite{wang2018non} significantly increases the number of parameters and FLOPs of the model. Interestingly, our CoResNet provides competitive results even when we compare it to the work of Hu et al.~\cite{hu2019squeeze}, although their work proposes an additional attention block (squeeze-and-excitation block) that needs to be inserted into the CNN, thus increasing the number of learnable parameters of the model. 

\subsection{Object detection on MS COCO}

In order to show the generality and the transfer learning capability of our approach, we integrate CoResNet in an object detection pipeline, namely  the Single Shot Multi-Box Detector (SSD)~\cite{liu2016ssd}. As in~\cite{liu2016ssd}, we remove all the layers of the ResNet backbones after the third stage to maintain the efficiency of the SSD. We also set the stride to 1 for the third stage to obtain $38\!\times\!38$ output feature maps from the backbones. The corresponding results, which are presented in Table~\ref{table:coco}, show that our approach provides significant improvements in detection performance, without affecting the number of parameters and computational cost. 

\subsection{Image generation on CIFAR-10 and CelebA}

In Table~\ref{table:image_generation}, we compare our CoProGAN with ProGAN \cite{karras-ICLR-2018} on CIFAR-10 and CelebA, respectively. While the IS on CIFAR-10 indicates that our model is slightly better, the FID points to a lager difference in favor of CoProGAN. Analogously, on CelebA, the MS-SSIM indicates slight performance gains brought by CoConv, but the improvements measured by SWD are much higher, regardless of the resolution of the output (from $16\!\times\!16$ to $128\!\times\!128$ pixels). Overall, the empirical evidence indicates that CoProGAN produces superior results, regardless of the metric. 

In Figure~\ref{fig:image_generation}, we show the best and worst looking examples generated by ProGAN and CoProGAN selected by an independent human annotator from a set of 50 images generated from each data set. On CIFAR-10, we observe that our successful results seem more realistic, while our failure cases seems to contain objects with a global structure. On CelebA, our successful faces seem more symmetrical, while the faces seen in the CoProGAN failure cases are still distinguishable as faces.

\section{Conclusion}

We proposed contextual convolution (CoConv) as a direct replacement of the standard convolution, aiming to integrate contextual information at different levels of neural architectures. CoConv is efficient, maintaining the same requirements in the number of parameters and computational costs as the standard convolution, while providing improved visual recognition capabilities. 
Our contextual convolutional neural network (CoCNN) architectures are motivated by a series of neuroscience studies which clearly indicate the presence and importance of contextual modulation, even at the early stages of the biological visual systems, specifically in the V1 area. In this work, we showed that the findings in neuroscience can be applied to the artificial visual systems for object detection, recognition and generation, where we obtain significant improvements over several state-of-the-art baselines.

\vspace{-0.1cm}
\section*{Acknowledgments}
\vspace{-0.1cm}
This work was supported by a grant of the Romanian Ministry of Education and Research, CNCS - UEFISCDI, project number PN-III-P1-1.1-TE-2019-0235, within PNCDI III. This article has also benefited from the support of the Romanian Young Academy, which is funded by Stiftung Mercator and the Alexander von Humboldt Foundation for the period 2020-2022.

{\small
\bibliographystyle{ieee_fullname}
\bibliography{references}

\begin{thebibliography}{10}\itemsep=-1pt

\bibitem{albright2002contextual}
Thomas~D Albright and Gene~R Stoner.
\newblock Contextual influences on visual processing.
\newblock {\em Annual Review of Neuroscience}, 25(1):339--379, 2002.

\bibitem{chen2017deeplab}
Liang-Chieh Chen, George Papandreou, Iasonas Kokkinos, Kevin Murphy, and Alan~L
  Yuille.
\newblock {DeepLab: Semantic Image Segmentation with Deep Convolutional Nets,
  Atrous Convolution, and Fully Connected CRFs}.
\newblock {\em IEEE Transactions on Pattern Analysis and Machine Intelligence},
  40(4):834--848, 2018.

\bibitem{chollet2017xception}
Fran{\c{c}}ois Chollet.
\newblock Xception: Deep learning with depthwise separable convolutions.
\newblock In {\em Proceedings of CVPR}, pages 1251--1258, 2017.

\bibitem{gilbert1990influence}
Charles~D Gilbert and Torsten~N Wiesel.
\newblock The influence of contextual stimuli on the orientation selectivity of
  cells in primary visual cortex of the cat.
\newblock {\em Vision Research}, 30(11):1689--1701, 1990.

\bibitem{Goodfellow-NIPS-2014}
Ian Goodfellow, Jean Pouget-Abadie, Mehdi Mirza, Bing Xu, David Warde-Farley,
  Sherjil Ozair, Aaron Courville, and Yoshua Bengio.
\newblock Generative adversarial nets.
\newblock In {\em Proceedings of NIPS}, pages 2672--2680, 2014.

\bibitem{goyal2017accurate}
Priya Goyal, Piotr Doll{\'a}r, Ross Girshick, Pieter Noordhuis, Lukasz
  Wesolowski, Aapo Kyrola, Andrew Tulloch, Yangqing Jia, and Kaiming He.
\newblock {Accurate, Large Minibatch SGD: Training ImageNet in 1 Hour}.
\newblock {\em arXiv:1706.02677}, 2017.

\bibitem{he2017mask}
Kaiming He, Georgia Gkioxari, Piotr Doll{\'a}r, and Ross Girshick.
\newblock {Mask R-CNN}.
\newblock In {\em Proceedings of ICCV}, pages 2961--2969, 2017.

\bibitem{he2016deep}
Kaiming He, Xiangyu Zhang, Shaoqing Ren, and Jian Sun.
\newblock {Deep Residual Learning for Image Recognition}.
\newblock In {\em Proceedings of CVPR}, pages 770--778, 2016.

\bibitem{he2016identity}
Kaiming He, Xiangyu Zhang, Shaoqing Ren, and J. Sun.
\newblock {Identity Mappings in Deep Residual Networks}.
\newblock In {\em Proceedings of ECCV}, pages 630--645, 2016.

\bibitem{Heusel-NIPS-2017}
Martin Heusel, Hubert Ramsauer, Thomas Unterthiner, Bernhard Nessler, and Sepp
  Hochreiter.
\newblock {GANs Trained by a Two Time-Scale Update Rule Converge to a Local
  Nash Equilibrium}.
\newblock In {\em Proceedings of NIPS}, pages 6626--6637, 2017.

\bibitem{hu2019squeeze}
Jie Hu, Li Shen, Samuel Albanie, Gang Sun, and Enhua Wu.
\newblock {Squeeze-and-Excitation Networks}.
\newblock {\em IEEE Transactions on Pattern Analysis and Machine Intelligence},
  42(8):2011--2023, 2020.

\bibitem{huang2017densely}
Gao Huang, Zhuang Liu, Laurens Van Der~Maaten, and Kilian~Q Weinberger.
\newblock {Densely Connected Convolutional Networks}.
\newblock In {\em Proceedings of CVPR}, pages 4700--4708, 2017.

\bibitem{ioffe2015batch}
Sergey Ioffe and Christian Szegedy.
\newblock {Batch Normalization: Accelerating Deep Network Training by Reducing
  Internal Covariate Shift}.
\newblock In {\em Proceedings of ICML}, pages 448--456, 2015.

\bibitem{karras-ICLR-2018}
Tero Karras, Timo Aila, Samuli Laine, and Jaakko Lehtinen.
\newblock {Progressive Growing of {GAN}s for Improved Quality, Stability, and
  Variation}.
\newblock In {\em Proceedings of ICLR}, 2018.

\bibitem{Kingma-ICLR-1015}
Diederik~P Kingma and Jimmy~Lei Ba.
\newblock Adam: A method for stochastic gradient descent.
\newblock In {\em Proceedings of ICLR}, 2015.

\bibitem{Krizhevsky-TR-2009}
Alex Krizhevsky and Geoffrey Hinton.
\newblock Learning multiple layers of features from tiny images.
\newblock Technical report, University of Toronto, 2009.

\bibitem{krizhevsky2012imagenet}
Alex Krizhevsky, Ilya Sutskever, and Geoffrey~E. Hinton.
\newblock {ImageNet classification with deep convolutional neural networks}.
\newblock In {\em Proceedings of NIPS}, pages 1097--1105, 2012.

\bibitem{lecun1989backpropagation}
Yann LeCun, Bernhard Boser, John~S Denker, Donnie Henderson, Richard~E Howard,
  Wayne Hubbard, and Lawrence~D Jackel.
\newblock Backpropagation applied to handwritten zip code recognition.
\newblock {\em Neural Computation}, 1(4):541--551, 1989.

\bibitem{lecun1998gradient}
Yann LeCun, L{\'e}on Bottou, Yoshua Bengio, Patrick Haffner, et~al.
\newblock Gradient-based learning applied to document recognition.
\newblock {\em Proceedings of the IEEE}, 86(11):2278--2324, 1998.

\bibitem{lin2017feature}
Tsung-Yi Lin, Piotr Doll{\'a}r, Ross Girshick, Kaiming He, Bharath Hariharan,
  and Serge Belongie.
\newblock {Feature Pyramid Networks for Object Detection}.
\newblock In {\em Proceedings of CVPR}, pages 2117--2125, 2017.

\bibitem{lin2017focal}
Tsung-Yi Lin, Priya Goyal, Ross Girshick, Kaiming He, and Piotr Doll{\'a}r.
\newblock {Focal Loss for Dense Object Detection}.
\newblock In {\em Proceedings of ICCV}, pages 2980--2988, 2017.

\bibitem{lin2014microsoft}
Tsung-Yi Lin, Michael Maire, Serge Belongie, James Hays, Pietro Perona, Deva
  Ramanan, Piotr Doll{\'a}r, and C~Lawrence Zitnick.
\newblock {Microsoft COCO: Common Objects in Context}.
\newblock In {\em {Proceedings of ECCV}}, pages 740--755, 2014.

\bibitem{liu2016ssd}
Wei Liu, Dragomir Anguelov, Dumitru Erhan, Christian Szegedy, Scott Reed,
  Cheng-Yang Fu, and Alexander~C Berg.
\newblock {SSD: Single Shot Multibox Detector}.
\newblock In {\em Proceedings of ECCV}, pages 21--37, 2016.

\bibitem{Liu-ICCV-2015}
Ziwei Liu, Ping Luo, Xiaogang Wang, and Xiaoou Tang.
\newblock {Deep Learning Face Attributes in the Wild}.
\newblock In {\em Proceedings of ICCV}, pages 3730--3738, 2015.

\bibitem{nair2010rectified}
Vinod Nair and Geoffrey~E. Hinton.
\newblock {Rectified Linear Units Improve Restricted Boltzmann Machines}.
\newblock In {\em Proceedings of ICML}, pages 807--814, 2010.

\bibitem{Rabin-SSVMCV-2012}
Julien Rabin, Gabriel Peyr{\'e}, Julie Delon, and Marc Bernot.
\newblock {Wasserstein Barycenter and Its Application to Texture Mixing}.
\newblock In {\em Proceedings of SSVM}, pages 435--446, 2012.

\bibitem{russakovsky2015imagenet}
Olga Russakovsky, Jia Deng, Hao Su, Jonathan Krause, Sanjeev Satheesh, Sean Ma,
  Zhiheng Huang, Andrej Karpathy, Aditya Khosla, Michael Bernstein, et~al.
\newblock {ImageNet Large Scale Visual Recognition Challenge}.
\newblock {\em International Journal of Computer Vision}, 115(3):211--252,
  2015.

\bibitem{sabour2017dynamic}
Sara Sabour, Nicholas Frosst, and Geoffrey~E. Hinton.
\newblock Dynamic routing between capsules.
\newblock In {\em Proceedings of NIPS}, pages 3856--3866, 2017.

\bibitem{Salimans-NIPS-2016}
Tim Salimans, Ian Goodfellow, Wojciech Zaremba, Vicki Cheung, Alec Radford, and
  Xi Chen.
\newblock {Improved techniques for training GANs}.
\newblock In {\em Proceedings of NIPS}, pages 2234--2242, 2016.

\bibitem{simonyan2014very}
Karen Simonyan and Andrew Zisserman.
\newblock {Very Deep Convolutional Networks for Large-Scale Image Recognition}.
\newblock {\em arXiv:1409.1556}, 2014.

\bibitem{szegedy2015going}
Christian Szegedy, Wei Liu, Yangqing Jia, Pierre Sermanet, Scott Reed, Dragomir
  Anguelov, Dumitru Erhan, Vincent Vanhoucke, and Andrew Rabinovich.
\newblock Going deeper with convolutions.
\newblock In {\em Proceedings of CVPR}, pages 1--9, 2015.

\bibitem{wang2018non}
Xiaolong Wang, Ross Girshick, Abhinav Gupta, and Kaiming He.
\newblock {Non-local Neural Networks}.
\newblock In {\em Proceedings of CVPR}, pages 7794--7803, 2018.

\bibitem{Wang-ACSSC-2003}
Z. Wang, E.P. Simoncelli, and A.C. Bovik.
\newblock Multiscale structural similarity for image quality assessment.
\newblock In {\em Proceedings of ACSSC}, volume~2, pages 1398--1402, 2003.

\bibitem{wu2018group}
Yuxin Wu and Kaiming He.
\newblock Group normalization.
\newblock In {\em Proceedings of ECCV}, pages 3--19, 2018.

\bibitem{xie2017aggregated}
Saining Xie, Ross Girshick, Piotr Doll{\'a}r, Zhuowen Tu, and Kaiming He.
\newblock {Aggregated Residual Transformations for Deep Neural Networks}.
\newblock In {\em Proceedings of CVPR}, pages 1492--1500, 2017.

\bibitem{yu2015multi}
Fisher Yu and Vladlen Koltun.
\newblock {Multi-Scale Context Aggregation by Dilated Convolutions}.
\newblock In {\em Proceedings of ICLR}, 2016.

\bibitem{yu2017dilated}
Fisher Yu, Vladlen Koltun, and Thomas Funkhouser.
\newblock {Dilated Residual Networks}.
\newblock In {\em Proceedings of CVPR}, pages 472--480, 2017.

\bibitem{zipser1996contextual}
Karl Zipser, Victor~AF Lamme, and Peter~H Schiller.
\newblock Contextual modulation in primary visual cortex.
\newblock {\em Journal of Neuroscience}, 16(22):7376--7389, 1996.

\bibitem{zoph2018learning}
Barret Zoph, Vijay Vasudevan, Jonathon Shlens, and Quoc~V Le.
\newblock {Learning Transferable Architectures for Scalable Image Recognition}.
\newblock In {\em Proceedings of CVPR}, pages 8697--8710, 2018.

\end{thebibliography}
}

\end{document}